\documentclass{article}
\usepackage{spconf,amsmath,graphicx}

\title{ASR RESCORING AND CONFIDENCE ESTIMATION WITH ELECTRA}
\name{Hayato Futami, Hirofumi Inaguma, Masato Mimura, Shinsuke Sakai, Tatsuya Kawahara}
\address{Graduate School of Informatics, Kyoto University, Sakyo-ku, Kyoto, Japan}

\usepackage{bm}
\usepackage{url}

\begin{document}
\ninept
\maketitle

\begin{abstract}
In automatic speech recognition (ASR) rescoring, the hypothesis with the fewest errors should be selected from the $n$-best list using a language model (LM).
However, LMs are usually trained to maximize the likelihood of correct word sequences, not to detect ASR errors.
We propose an ASR rescoring method for directly detecting errors with ELECTRA, which is originally a pre-training method for NLP tasks.
ELECTRA is pre-trained to predict whether each word is replaced by BERT or not, which can simulate ASR error detection on large text corpora.
To make this pre-training closer to ASR error detection, we further propose an extended version of ELECTRA called phone-attentive ELECTRA (P-ELECTRA).
In the pre-training of P-ELECTRA, each word is replaced by a phone-to-word conversion model, which leverages phone information to generate acoustically similar words.
Since our rescoring method is optimized for detecting errors, it can also be used for word-level confidence estimation.
Experimental evaluations on the Librispeech and TED-LIUM2 corpora show that our rescoring method with ELECTRA is competitive with conventional rescoring methods with faster inference.
ELECTRA also performs better in confidence estimation than BERT because it can learn to detect inappropriate words not only in fine-tuning but also in pre-training.
\end{abstract}



\section{Introduction}
Automatic speech recognition (ASR) has been realized by DNN-HMM hybrid systems with an acoustic model (AM) and a language model (LM).
Recently, end-to-end ASR that integrates an AM and an LM into a single neural network has achieved prominent performances.
There are some architectures for end-to-end modeling: connectionist temporal classification (CTC) \cite{Graves06-CTC}, attention-based encoder-decoder models \cite{Chan16-LAS, Dong18-ST, Gulati20-CF}, and neural network transducer models \cite{Graves12-ST, Zhang20-TT}.

End-to-end ASR models are trained on paired speech and transcripts.
While the amount of paired data for the target domain is limited, a large amount of text-only data is often available.
An external LM trained on text-only data is usually applied to improve ASR performance.
Shallow Fusion \cite{Chorowski17-TBD, Irie19-LMDT, Kannan18-AIE} and $n$-best rescoring are widely used for applying an external LM to end-to-end ASR.
In Shallow Fusion, the interpolated score of the LM and the ASR model is used in beam search decoding.
In $n$-best rescoring, which we focus on in this study, an $n$-best list obtained from the ASR model is scored using the LM, then the hypothesis of the highest interpolated score is selected.

RNN and Transformer LMs are conventionally used for rescoring.
They predict each word on the basis of its preceding context in an autoregressive manner.
These autoregressive LMs are usually trained on sentences without errors by maximizing the likelihood of the word sequences.
In rescoring, however, LMs should discriminate a hypothesis with fewer ASR errors from other hypotheses with more errors.
Several studies have pointed out that the training objective of LMs is sub-optimal for rescoring, and they proposed discriminative LMs \cite{Tachioka15-DLM, Hori16-MWER, Huang18-LM}, which are trained with discriminative criteria using ASR results and their corresponding references.
They are obtained from paired speech and transcripts, the scale of which is usually limited compared with text-only data.

Recently, BERT \cite{Devlin19-BERT} has been used for rescoring \cite{Shin19-ESS, Salazar20-MLMS}.
BERT was originally proposed as a pre-training method for NLP tasks such as question answering and language understanding.
BERT can also be regarded as an external LM that predicts each masked word on the basis of both its preceding and following context.
BERT performs better in rescoring than conventional autoregressive LMs because of the deeply bidirectional architecture of its Transformer encoder \cite{Shin19-ESS, Salazar20-MLMS}.
However, rescoring with BERT is too time-consuming.
It takes $L$ inference steps to rescore a hypothesis of length $L$ \cite{Salazar20-MLMS} by masking each word iteratively, while Transformer LM takes a single step \cite{Li20-PR}.


In this study, we propose an ASR rescoring method for detecting then counting ASR errors with ELECTRA \cite{Clark20-ELECTRA}, which is a pre-training method with a deeply bidirectional architecture like BERT.
Different from BERT, ELECTRA is pre-trained for a replaced token detection task instead of masked language model (MLM), in which the generator (BERT) replaces some words of the input by sampling and the discriminator is trained to predict whether each word is replaced by BERT or not.
ASR error detection can be trained only on paired data, but ELECTRA can simulate it on large text corpora.
However, there is a mismatch between real ASR errors and BERT's replacement used in the pre-training of ELECTRA because ASR takes acoustic features as input, while BERT does not.
To solve the mismatch, we further investigate two methods of making ELECTRA's training conditions closer to rescoring conditions: fine-tuning ELECTRA on ASR results for error detection and introducing phone-attentive ELECTRA (P-ELECTRA).
P-ELECTRA is a modified version of ELECTRA, which employs a phone-to-word conversion model as a generator instead of BERT.
With the help of phone information, we can obtain replacements that are similar to ASR errors on text corpora.

Our rescoring method with ELECTRA can solve the two issues mentioned above: the mismatch between LM training and rescoring objectives and slow inference in rescoring with bidirectional architecture.
First, it is optimized directly to the rescoring objective by detecting errors in each hypothesis and selecting the hypothesis with the fewest errors in the $n$-best list.
Second, it can benefit from deeply bidirectional contextual information with faster inference than BERT, as it takes only a single step to rescore without masking.
Moreover, rescoring with ELECTRA is faster than with conventional LMs.
ELECTRA conducts binary classification for rescoring, while conventional LMs conduct word prediction, the complexity of which is proportional to the vocabulary size.

The proposed rescoring method is closely related to confidence estimation, or the ASR error detection task.
Confidence estimation assesses the quality of ASR predictions \cite{Wessel01-CM, Kawahara98-FSU, Ragni18-CEDP, Swarup19-ALEXA, Li21-CE, Alejandro20-CM, Ogawa21-BCE, Oneata21-EWCE}, which is useful for many downstream ASR applications such as voice assistants.
We demonstrate that our models for rescoring can be applied to confidence estimation without any additional architectural changes or training.
ELECTRA is pre-trained for the replaced token detection task that is close to confidence estimation, and therefore it can effectively leverage text-only data.


\section{Preliminaries and related work}

\subsection{ELECTRA}
ELECTRA \cite{Clark20-ELECTRA} is a pre-training method for downstream NLP tasks like BERT \cite{Devlin19-BERT}.
In BERT, masked language modeling (MLM) replaces some input tokens with \url{[MASK]}, then it is trained to predict the original tokens.
ELECTRA instead employs a replaced token detection task for pre-training.
It corrupts the input by replacing some tokens by sampling from a generator, then a discriminator is trained to predict whether each token is the original or replacement.
While BERT learns from a small masked-out subset (usually $15\%$), ELECTRA can learn from all input tokens, which is more computationally efficient \cite{Clark20-ELECTRA}.

In the pre-training of ELECTRA, two Transformer-based models (generator $G$ and discriminator $D$) are jointly trained.
The procedure is formulated as follows.
First, some tokens of the input $\bm{y} = (y_1, y_2, ..., y_L)$ are selected and masked out.
Let $\bm{m} = (m_1, m_2, ..., m_k) (1 \leq m_i \leq L)$ denote the positions of masked-out tokens.
\begin{align}
    \bm{y}^{\rm masked} = {\rm replace}(\bm{y}, \bm{m}, {\rm \hbox{\url{[MASK]}}})
\end{align}
The generator $G$ then predicts tokens of the masked positions and generates a corrupted example $\bm{y}^{\rm corrupt}$ by sampling.
\begin{align}
    \bm{y}^{\rm corrupt} = {\rm replace}(\bm{y}^{\rm masked}, \bm{m}, \hat{\bm{y}}) \nonumber \\
    \hat{y}_i \sim p_G(y_i | \bm{y}^{\rm masked}), i \in \bm{m}
\end{align}
The generator is BERT trained with the MLM objective, and its loss function is
\begin{align}
    \mathcal{L}_G = \sum_{i \in \bm{m}} - \log p_G(y_i | \bm{y}^{\rm masked})
\end{align}
The discriminator $D$ is a binary classifier trained to discriminate the original token from the token replaced by the generator $G$.
Let $D^{(i)}(\bm{y})$ denote the discriminator output passed through the sigmoid layer for the $i$-th token, which should be $1$ when $y_i$ is replaced.
Its loss function is
\begin{align}
\label{eq:loss-disc}
    \mathcal{L}_D = \sum_{i=1}^L - \delta(y^{\rm corrupt}_i, y_i) \log (1 - D^{(i)}(\bm{y}^{\rm corrupt})) \nonumber \\
    - (1 - \delta(y^{\rm corrupt}_i, y_i)) \log D^{(i)}(\bm{y}^{\rm corrupt})
\end{align}
where $\delta(y^{\rm corrupt}_i, y_i)$ becomes $1$ when $y^{\rm corrupt}_i = y_i$, and $0$ otherwise.
The weighted sum of $\mathcal{L}_G$ and $\mathcal{L}_D$ is minimized during pre-training.
After pre-training, the discriminator is fine-tuned on downstream NLP tasks, and it is referred to as ``ELECTRA''.
In this study, we use the discriminator for rescoring and confidence estimation tasks.

\subsection{Rescoring}
Rescoring is a simple and widely used method to apply LMs to end-to-end ASR models.
An $n$-best list is generated by beam search with the ASR model, then each hypothesis in the list is rescored using the LM.
\begin{align}
\label{eq:rescore}
    Score(\bm{X}, \bm{y}) = \log p_{\rm ASR}(\bm{y} | \bm{X}) + \alpha Score_{\rm LM}(\bm{y}) + \beta |\bm{y}|
\end{align}
where $\bm{X}$ denotes acoustic features, and $\bm{y} = (y_1, ..., y_L)$ denotes a hypothesis.
The hypothesis that has the highest $Score(\bm{X}, \bm{y})$ is selected.
As $\log p_{\rm ASR}$ and $Score_{\rm LM}$ tend to assign a higher score to a shorter hypothesis, $\beta |\bm{y}|$ encourages longer hypotheses to reduce deletion errors.
Here, $\alpha$ and $\beta$ are hyperparameters.

The $Score_{\rm LM}$ is conventionally calculated using autoregressive LMs such as RNN and Transformer LMs.
They provide a likelihood score for each hypothesis $\bm{y}$ as
\begin{align}
\label{eq:score-lm}
    Score_{\rm LM}(\bm{y}) = \sum_{i=1}^L \log p(y_i | \bm{y}_{<i}) = \log p(y_1, ..., y_L)
\end{align}
where $\bm{y}_{<i} = (y_1, ..., y_{i-1})$.
Autoregressive LMs predict each token using its preceding context.
Transformer LM can calculate $p(y_i | \bm{y}_{<i})$ for all $i$ in parallel with self-attention mechanism \cite{Li20-PR}.

Recently, BERT has also been used for calculating $Score_{\rm LM}$.
It provides a pseudo-likelihood score \cite{Wang19-BERT} as
\begin{align}
\label{eq:score-bert}
    Score_{\rm LM}(\bm{y}) = \sum_{i=1}^L \log p(y_i | \bm{y}_{\backslash i})
\end{align}
where $\bm{y}_{\backslash i} = (y_1, ..., y_{i-1}, $\url{[MASK]}$, y_{i+1}, ..., y_L)$ (the $i$-th token of $\bm{y}$ is masked out).
BERT is reported to perform better in rescoring than autoregressive LMs by predicting each token using both its preceding and following context \cite{Shin19-ESS, Salazar20-MLMS}.
It also performs better than bidirectional RNN LMs \cite{Arisoy15-BRNN, Chen17-IBRNN}, because it is ``deeply bidirectional'' \cite{Devlin19-BERT}, while the bidirectional RNN is the shallow concatenation of two directions of RNNs \cite{Peters18-DCWR}.
However, rescoring with BERT takes $L$ inference steps for a hypothesis of length $L$.
It requires a iterative procedure to calculate $p(y_i | \bm{y}_{\backslash i})$ for each position $i$ using different masked inputs $\bm{y}_{\backslash i}$.
More recently, Electric \cite{Clark20-PTE} was proposed to calculate $p(y_i | \bm{y}_{\backslash i})$ for all $i$ in a single step.
It efficiently provides $Score_{\rm LM}$ based on pseudo-likelihood through noise contrastive estimation training.
Electric only learns the training data distribution, while ELECTRA learns whether each token comes from the data distribution or noise distribution by the generator in pre-training \cite{Clark20-PTE}.
ELECTRA provides $Score_{\rm LM}$ based on error detection.
It can be fine-tuned on ASR hypotheses and directly applied to confidence estimation.

Another stream of ASR rescoring includes discriminative language modeling.
LMs are usually trained to maximize the likelihood of word sequences without errors.
However, this does not necessarily maximize the performance of ASR rescoring, in which the LM should discriminate the hypothesis with the fewest errors from hypotheses that contain more errors.
In \cite{Tachioka15-DLM}, the word-level log-likelihood ratio of ASR hypotheses and references was used as a training criterion for RNN LMs.
In \cite{Hori16-MWER, Gandhe20-AADLM}, a minimum word error rate (MWER) training for RNN LMs was proposed.
In \cite{Huang18-LM, Wang21-LM}, a large margin criterion that enlarges the margin between hypotheses and references was applied to RNN LM, Transformer LM, and BERT training.
In these studies, the likelihood of each word or sentence was adapted to the discriminative criterion using ASR hypotheses and references.
In this study, we use ELECTRA to learn error detection that is in nature discriminative and can be trained not only on paired data but also on large text corpora.

\subsection{Confidence estimation}
\label{sec:related-work-confidence-estimation}
Confidence estimation is an important task for ASR applications to mitigate the adverse effects of ASR errors, which are inevitable.
For example, in voice assistants, queries of low confidence will be asked back.
In audio transcription tasks, it helps manual corrections by flagging less confident words.
Confidence estimation is also used in semi-supervised learning \cite{Park20-INST}, in which utterances with confident predictions are selected as training data.

Confidence scores are estimated at word-level or utterance-level.
In this study, we focus on word-level confidence estimation. 
In DNN-HMM ASR, reliable word-level confidence scores can be obtained from word posterior probabilities over lattices \cite{Wessel01-CM}, and they are further improved by confidence estimation modules (CEMs) \cite{Kawahara98-FSU, Ragni18-CEDP, Swarup19-ALEXA}.
In seq2seq ASR models, confidence scores can be obtained from softmax probabilities of their decoders, but they are not reliable enough because of overconfidence \cite{Guo17-OC}.  
In \cite{Li21-CE}, a lightweight CEM that uses internal features of a seq2seq model was proposed to mitigate overconfidence.
In \cite{Alejandro20-CM}, softmax temperature values for each token were predicted to adjust overconfident probabilities.
In CTC-based ASR models we used in this study, confidence scores can be obtained with the forward-backward algorithm \cite{Watanabe17-HCA}, which was reported to perform well \cite{Ogawa21-BCE}.
End-to-end ASR models and CEMs usually use subword units such as byte pair encoding (BPE) \cite{sennrich16-BPE} to handle rare and unknown words.
They provide subword-level confidence scores, but in many downstream applications, word-level confidence scores are useful.
Word-level confidence scores can be obtained simply by taking the average \cite{Li21-CE}, product, or minimum \cite{Oneata21-EWCE} of subword-level scores if a word consists of multiple subwords.
In \cite{Qiu21-LWLC}, a subword-level CEM was directly trained with the word-level confidence objective.
In these studies, CEMs were mainly trained with ASR results and their references obtained from paired data.
In this study, ELECTRA is used as a CEM, which can be pre-trained by simulating the confidence estimation task on large text corpora.

\section{Proposed method}

\subsection{Rescoring with ELECTRA}
\begin{figure}[t]
  \centering
  \includegraphics[width=\linewidth]{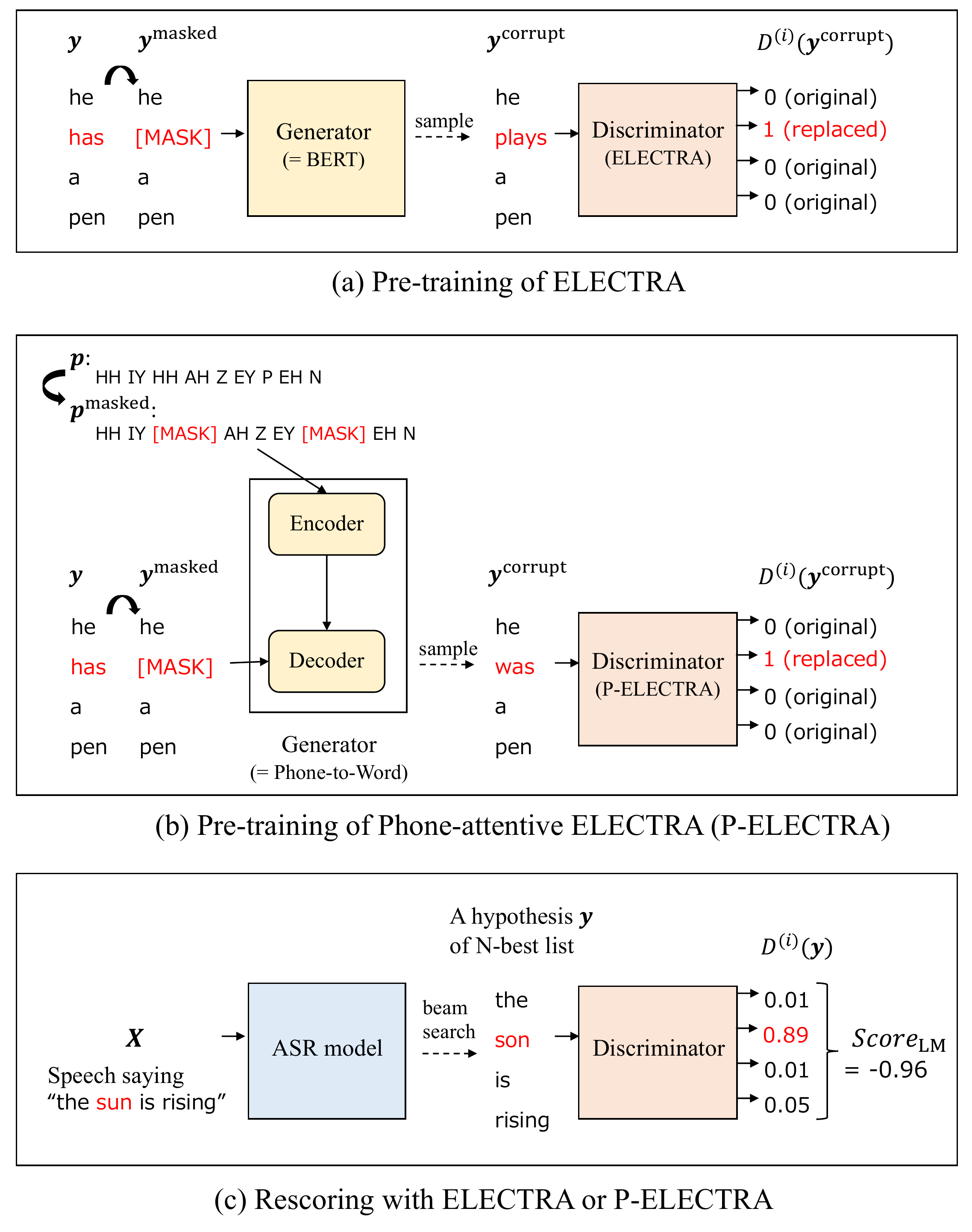}
  \caption{Overview of our rescoring method. First, ELECTRA or P-ELECTRA is pre-trained on text data (a) or (b). It can then be fine-tuned on ASR hypotheses for error detection. Finally, it assigns $Score_{\rm LM}$ to each hypothesis for rescoring (c). ASR model can be regarded as generator that produces corrupted input.}
  \label{fig:overview}
\end{figure}

In ELECTRA, $D^{(i)}(\bm{y})$ is pre-trained to output 1 when the $i$-th token is replaced, and 0 otherwise, as illustrated in Fig. \ref{fig:overview} (a).
ELECTRA learns to detect syntactically or semantically inappropriate tokens, which is useful for detecting ASR errors.
By counting the expected number of errors in a hypothesis, $Score_{\rm LM}(\bm{y})$ is defined as
\begin{align}
\label{eq:score-electra}
Score_{\rm LM}(\bm{y}) = - \sum_{i=1}^L D^{(i)}(\bm{y})
\end{align}
which is illustrated in Fig. \ref{fig:overview} (c).
Note that the score should be higher for hypotheses with fewer errors.
Rescoring with ELECTRA can benefit from deeply bidirectional contextual information with a single-step inference per hypothesis, while BERT requires $L$ steps.
ELECTRA can look at all the tokens including $y_i$ to calculate $D^{(i)}(\bm{y})$, and therefore it provides $D^{(i)}(\bm{y})$ for all $i$ in parallel. 
Even compared with Transformer LM, ELECTRA is faster because it has a sigmoid layer for outputs.
Transformer LM has a softmax layer, the computation of which is proportional to the vocabulary size.

\subsection{Fine-tuning on ASR hypotheses}
In pre-training, ELECTRA is trained to detect samples generated from BERT.
However, in rescoring, ELECTRA needs to detect errors generated from the ASR model.
BERT generates tokens on the basis of linguistic context, but the ASR model generates tokens on the basis of both acoustic features and linguistic context.
Acoustically similar tokens, such as ``two'' and ``too'', tend to be mistaken in ASR.
To solve the mismatch between pre-training and rescoring conditions, we fine-tune ELECTRA on ASR hypotheses including real ASR errors.
The ASR hypotheses are aligned with their corresponding references and each word of them is labeled as correct or incorrect.
They can be obtained from paired speech and transcripts used for ASR training.
Note that only the discriminator is fine-tuned.

\begingroup
\renewcommand{\arraystretch}{1.1}
\begin{table*}[t]
  \caption{Examples of generated tokens in pre-training of ELECTRA and P-ELECTRA. In addition to $\bm{y}^{\rm masked}$, phone tokens are used as inputs in pre-training of P-ELECTRA. ``\_'' denotes word boundary.}
  \vspace{10pt}
  \label{tab:example-generation}
  \centering
  \begin{tabular}{ll} \hline
    $\bm{y}$: & \_i \_don ' t \_believe \textbf{\_ann} \textbf{\_knew} \_any \_magic \_or \_she ' d \_have \textbf{\_worked} \_it \_before \\
    $\bm{y}^{\rm masked}$: & \_i \_don ' t \_believe \url{[MASK]} \url{[MASK]} \_any \_magic \_or \_she ' d \_have \url{[MASK]} \_it \_before \\
    $\bm{y}^{\rm corrupt}$ {\small in ELECTRA}: & \_i \_don ' t \_believe  \textbf{\_there} \textbf{\_is} \_any \_magic \_or \_she ' d \_have \textbf{\_used} \_it \_before \\
    $\bm{y}^{\rm corrupt}$ {\small in P-ELECTRA}: & \_i \_don ' t \_believe \textbf{\_and} \textbf{\_you} \_any \_magic \_or \_she ' d \_have \textbf{\_worked} \_it \_before \\ \hline
 \end{tabular}
\end{table*}
\endgroup

\subsection{Phone-attentive ELECTRA}
\label{sec:p-electra}
Fine-tuning on ASR hypotheses can be done only on a limited amount of paired data, and therefore it is prone to overfitting.
To generate acoustically similar errors on text data, we leverage phone information that can be automatically obtained from word sequences using a pronunciation dictionary \cite{Masumura20-PGC, Tang21-GMT}.
We propose a modified version of ELECTRA called phone-attentive ELECTRA (P-ELECTRA).
In P-ELECTRA, the generator $G$ is a phone-to-word conversion model instead of BERT used in original ELECTRA.
It is a Transformer-based conditional masked language model (CMLM) \cite{Ghazvininejad19-MP}, which consists of an encoder and a decoder, as illustrated in Fig. \ref{fig:overview} (b).
Some tokens of phones are randomly replaced with \url{[MASK]} and fed into the encoder as $\bm{p}^{\rm masked}$ to prevent overfitting, which is called ``textaugment'' \cite{Wang21-CRNNT}.
Some tokens of words are also masked out and fed into the decoder as $\bm{y}^{\rm masked}$.
The generator predicts word tokens of the masked positions on the basis of both phone and word tokens and generates a corrupted example $\bm{y}^{\rm corrupt}$.
\begin{align}
\bm{y}^{\rm corrupt} = {\rm replace}(\bm{y}^{\rm masked}, \bm{m}, \hat{\bm{y}}) \nonumber \\
\hat{y}_i \sim p_{G}(y_i | \bm{y}^{\rm masked}, \bm{p}^{\rm masked}), i \in \bm{m}
\end{align}
Note that phones are not required in rescoring because only the discriminator is used.
For efficient training, sequences of phones and those of corresponding words are concatenated up to a certain length, respectively, then input to the generator.

Table \ref{tab:example-generation} compares the generated tokens by BERT in the pre-training of ELECTRA and those by the phone-to-word conversion model in the pre-training of P-ELECTRA.
In P-ELECTRA, ``\_ann \_knew'' is replaced with ``\_and \_you'' by using the phone-to-word conversion model, which is more acoustically similar and likely to appear in ASR results than the replacement by using BERT ``\_there \_is''.

\subsection{Confidence estimation with ELECTRA}
In ELECTRA (including P-ELECTRA), $c^{(i)}(\bm{y}) = 1 - D^{(i)}(\bm{y})$ can be regarded as a token-level confidence score.
When we use subword units such as BPE, a word-level confidence score $c_{\rm word}^{(i)}(\bm{y})$ is obtained by taking the minimum value of token-level scores for each word.
ELECTRA leverages the knowledge of large text corpora for confidence estimation.
BERT can also be applied to confidence estimation by pre-training for MLM then fine-tuning on ASR hypotheses.
For fine-tuning, a new sigmoid layer for outputs is added to BERT, and its parameters are trained from scratch.
On the other hand, ELECTRA has the sigmoid layer and can learn confidence scores even from the pre-training stage by detecting replaced tokens.
ELECTRA is also more robust against ASR errors, as it is pre-trained not on ground truth text but on corrupted examples, which is indicated in punctuation restoration \cite{Hentschel21-PR}.

As mentioned in Section \ref{sec:related-work-confidence-estimation}, the ASR model itself provides word-level confidence scores $p_{\rm word}^{(i)}(\bm{y} | \bm{X})$.
We interpolate the scores of ASR and ELECTRA as
\begin{align}
\label{eq:score-confidence-interpolate}
c'^{(i)}_{\rm word} = (1 - \gamma) p_{\rm word}^{(i)}(\bm{y} | \bm{X}) + \gamma c^{(i)}_{\rm word}(\bm{y})
\end{align}
where $\gamma$ is a tunable weight parameter.

\section{Experimental evaluations}

\subsection{Experimental conditions}
We evaluated our method using the Librispeech \cite{Libri15} and TED-LIUM2 \cite{Ted214} corpora.
The training data of Librispeech consists of $960$ hours of paired speech and transcripts, and about $800$ million words of additional text data.
Its evaluation data consist of a ``clean'' set with lower-WER speakers and ``other'' set with higher-WER ones.
The text is tokenized using BPE \cite{sennrich16-BPE} of vocabulary size $9951$ to make subword tokens.
It is also converted to phone tokens of vocabulary size $74$ using a lexicon.
The training data of TED-LIUM2 consist of $207$ hours of paired data and about $250$ million words of additional text.
The text is converted to subword tokens of vocabulary size $9798$ and phone tokens of vocabulary size $44$.

We prepared a CTC-based model for end-to-end ASR that consists of Transformer encoder with $12$ layers, $256$ hidden units, and $4$ attention heads.
We used the Adam \cite{Kingma15-Adam} optimizer with Noam learning rate scheduling \cite{Dong18-ST} of $warmup\_n = 25000, k = 5$.
SpecAugment \cite{Park19-SA} was applied to acoustic features.
Speed perturbation \cite{Ko15-AA} was also applied in the TED-LIUM2 experiments.
We obtained a $50$-best list with beam search decoding.

We prepared Transformer LM, BERT, ELECTRA, P-ELECTRA for rescoring and confidence estimation.
Transformer LM, BERT, the generator for ELECTRA and ELECTRA (discriminator), and P-ELECTRA (discriminator) consist of Transformer with $12$ layers, $256$ hidden units, and $4$ attention heads.
The generator for P-ELECTRA (a phone-to-word conversion model) consists of $4$-layer Transformer encoder and $4$-layer Transformer decoder, which has almost the same number of parameters as those of other models.
They are implemented with ``transformers'' library \cite{Wolf20-T}.
We pre-trained them on the Librispeech text data in Librispeech experiments and pre-trained them on the TED-LIUM2 text data in TED-LIUM2 experiments.
Note that only discriminators are used for rescoring and confidence estimation tasks in ELECTRA and P-ELECTRA.
We used the Adam optimizer with linear learning rate scheduling, in which the learning rate increases linearly for the first $10\%$ of the total steps to $0.0001$, thereafter decreasing linearly.
During pre-training, $15\%$ of the input subword tokens were selected and replaced with \url{[MASK]}.
Next sentence prediction \cite{Devlin19-BERT} is omitted from the pre-training objective of BERT in this study as in \cite{Liu19-RoBERTa}.
In the pre-training of P-ELECTRA, $30\%$ of the phone tokens were also replaced with \url{[MASK]}, as mentioned in Section \ref{sec:p-electra}.
We further fine-tuned BERT, ELECTRA, and P-ELECTRA for ASR error detection, or confidence estimation, using the $5$-best list generated from the ASR training data.
These fine-tuned models are denoted as BERT(FT), ELECTRA(FT), and P-ELECTRA(FT), respectively.

\subsection{Experimental results}

\begingroup
\renewcommand{\arraystretch}{1.1}
\begin{table}[t]
  \caption{Rescoring results on Librispeech. $x$(FT) denotes that model $x$ was fine-tuned on ASR $5$-best list. ``Runtime'' denotes runtime compared with Transformer LM. Bottom $4$ rows are results of proposed method.}
  \vspace{10pt}
  \label{tab:rescoring}
  \centering
  \begin{tabular}{lccc}  \hline
     & \multicolumn{2}{c}{WER(\%) $\downarrow$} & Runtime $\downarrow$ \\
     & clean & other & clean \\ \hline
    baseline ASR (CTC) & $5.80$ & $13.46$ & - \\
    +Transformer LM & $4.08$ & $10.45$ & $\times 1.0$ \\
    +BERT & $4.02$ & $\textbf{10.25}$ & $\times 27.1$ \\
    +BERT (FT) & $4.29$ & $10.71$ & $\times 0.8$ \\
    +ELECTRA & $4.07$ & $10.87$ & $\times 0.8$ \\
    +ELECTRA (FT) & $\textbf{3.98}$ & $10.53$ & $\times 0.8$ \\
    +P-ELECTRA & $4.04$ & $10.44$ & $\times 0.8$ \\
    +P-ELECTRA (FT) & $4.00$ & $10.42$ & $\times 0.8$ \\  \hline
    oracle & $2.47$ & $7.85$ & - \\ \hline
 \end{tabular}
\end{table}
\endgroup

\begin{figure*}[t]
\caption{Scoring comparison with Transformer LM and fine-tuned P-ELECTRA for Librispeech ``clean'' set. 
``$\times$'' denotes average $-Score_{\rm LM}$ for each ``Number of errors''.}
\label{fig:score-analysis}
\vspace{10pt}
\includegraphics[width=0.9\linewidth]{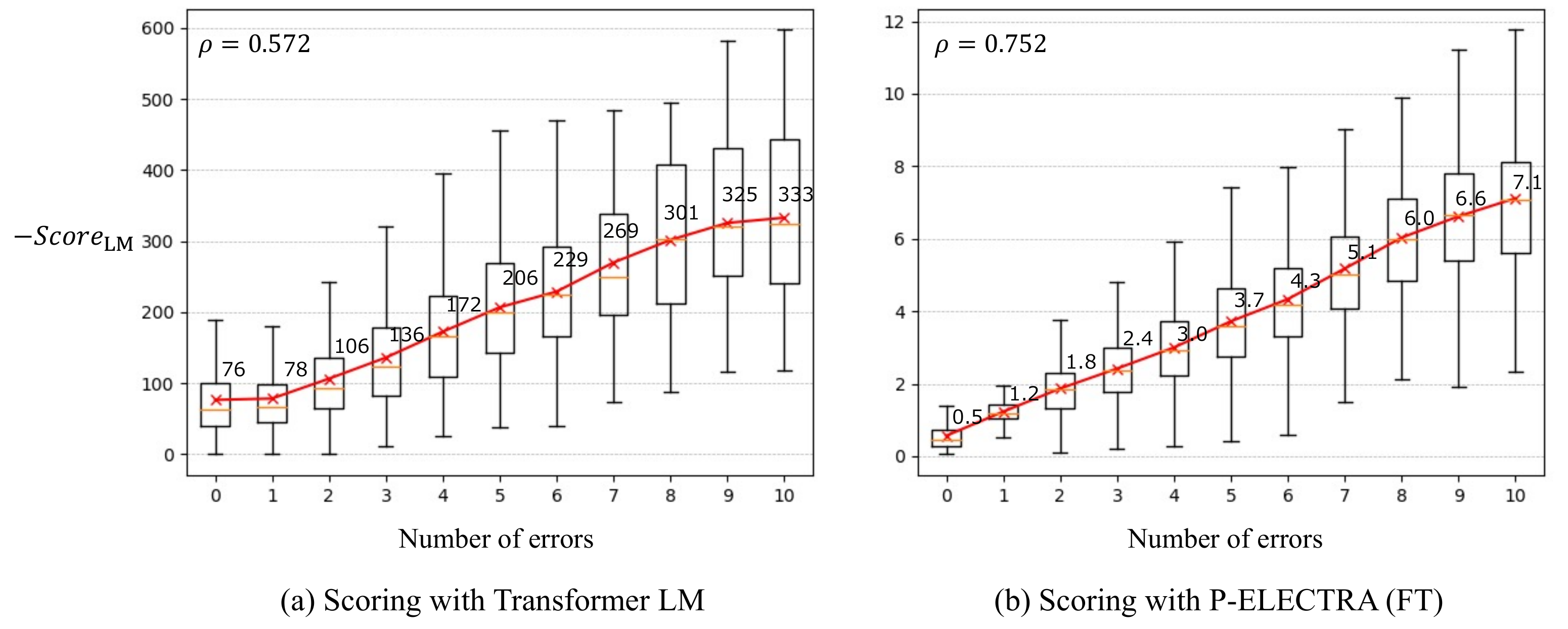}
\end{figure*}

\subsubsection{Rescoring on Librispeech}
Table \ref{tab:rescoring} compares rescoring results with different models on Librispeech.
Transformer LM and BERT provide $Score_{\rm LM}$ based on likelihood, as in Eqs. (\ref{eq:score-lm}) and (\ref{eq:score-bert}), respectively, and ELECTRAs (ELECTRA and P-ELECTRA) provide $Score_{\rm LM}$ based on error counting, as in Eq. (\ref{eq:score-electra}).
$\alpha$ and $\beta$ in Eq. (\ref{eq:rescore}) were adjusted using development sets.
As the range of log-likelihood and that of error counts differ, $\alpha$ for Transformer LM or BERT and that for ELECTRAs were also different.
For example, $\alpha = 0.5, \beta = 1.0$ were suitable for Transformer LM and $\alpha = 9.0, \beta = 0$ for ELECTRA.
In terms of word error rate (WER), BERT outperformed Transformer LM, which is consistent with previous studies \cite{Shin19-ESS, Salazar20-MLMS}.
ELECTRA pre-trained for replaced token detection detected ASR errors without fine-tuning and improved WER over the baseline ASR.
By fine-tuning ELECTRA on ASR hypotheses ($5$-best), it outperformed Transformer LM for the ``clean'' set.
Rescoring with P-ELECTRA outperformed ELECTRA and was competitive with Transformer LM without fine-tuning not only for the ``clean'' set but also for the ``other'' set.
We saw that the pre-training method using phones in P-ELECTRA was effective, especially for the ``other'' set that contains more ASR errors.
However, fine-tuning P-ELECTRA did not show much WER improvement.
$Score_{\rm LM}$ based on error counting in Eq. (\ref{eq:score-electra}) can also be calculated with fine-tuned BERT, but this was not as successful as with ELECTRAs.
We saw that the replaced token detection task for the pre-training of ELECTRAs is closer to the target rescoring task than MLM for BERT and thus lead to better rescoring.
Our rescoring with ELECTRAs requires only $80\%$ of the runtime of that with Transformer LM, as ELECTRAs do not compute the softmax layer of $V$ entries.
Rescoring with BERT resulted in the largest WER reduction especially for the ``other'' set, but its runtime was much inferior to the others, i.e. about $27$ times slower than Transformer LM.
It is dependent on not only $V$ but also the length of a hypothesis $L$ for each hypothesis.

Next, we compared $Score_{\rm LM}$ of Transformer LM and that of fine-tuned P-ELECTRA for the Librispeech ``clean'' set, as shown in Fig. \ref{fig:score-analysis}.
``Number of errors'' in the figure indicates {\it word-level} substitution, insertion, and deletion errors of each hypothesis against its reference, which are used in WER calculation.
$-Score_{\rm LM}$ of P-ELECTRA corresponds to the expected number of {\it subword-level} substitution and insertion errors.
It is correlated to the number of all types of {\it word-level} errors.
In P-ELECTRA, the average $-Score_{\rm LM}$ increases almost linearly with the number of errors.
Table \ref{tab:score-analysis} lists the correlation coefficients between $-Score_{\rm LM}$ and the actual number of errors.
Fine-tuning on actual ASR errors made the correlation higher.
P-ELECTRA achieved a high correlation with and even without fine-tuning, which indicates its pre-training using phone information simulated ASR error detection well.

\begingroup
\renewcommand{\arraystretch}{1.1}
\begin{table}[t]
  \caption{Correlation coefficients between LM scoring and number of errors. Scoring of Transformer LM and BERT is derived from likelihood (pseudo-likelihood), and scoring of bottom $5$ methods is derived from error counting.}
  \vspace{10pt}
  \label{tab:score-analysis}
  \centering
  \begin{tabular}{lcc} \hline
       & \multicolumn{2}{c}{Coefficients $\rho$} \\
       & clean & other \\ \hline
      Transformer LM & $0.572$ & $0.625$ \\
      BERT & $0.655$ & $0.728$ \\
      BERT (FT) & $0.716$ & $0.738$ \\
      ELECTRA & $0.620$ & $0.634$ \\
      ELECTRA (FT) & $0.740$ & $0.762$ \\
      P-ELECTRA & $0.687$ & $0.767$ \\
      P-ELECTRA (FT) & $\textbf{0.752}$ & $\textbf{0.788}$ \\ \hline
  \end{tabular}
\end{table}
\endgroup

\begingroup
\renewcommand{\arraystretch}{1.1}
\begin{table}[t]
  \caption{Confidence estimation results on Librispeech. $x$(FT) denotes that model (CEM) $x$ was fine-tuned on ASR $5$-best list. ASR+$x$ denotes confidence score interpolation between ASR and CEM $x$.}
  \vspace{10pt}
  \label{tab:confidence}
  \centering
  \begin{tabular}{lcccc} \hline
     & \multicolumn{2}{c}{AUC $\uparrow$} & \multicolumn{2}{c}{NCE $\uparrow$} \\
     & clean & other & clean & other \\ \hline
    ASR & $0.955$ & $0.927$ & $0.546$ & $0.426$ \\
    BERT(FT) & $0.945$ & $0.914$ & $0.544$ & $0.394$ \\
    ELECTRA & $0.903$ & $0.871$ & $0.418$ & $0.318$ \\
    ELECTRA(FT) & $0.956$ & $0.928$ & $0.594$ & $0.455$ \\
    P-ELECTRA & $0.922$ & $0.907$ & $0.472$ & $0.375$ \\
    P-ELECTRA(FT) & $\textbf{0.958}$ & $\textbf{0.935}$ & $\textbf{0.615}$ & $\textbf{0.499}$ \\ \hline
    ASR+BERT(FT) & $0.973$ & $0.952$ & $0.660$ & $0.535$ \\
    ASR+ELECTRA & $0.965$ & $0.942$ & $0.623$ & $0.499$  \\
    ASR+ELECTRA(FT) & $0.976$ & $0.956$ & $0.681$ & $0.560$ \\
    ASR+P-ELECTRA & $0.969$ & $0.951$ & $0.649$ & $0.541$ \\
    ASR+P-ELECTRA(FT) & $\textbf{0.977}$ & $\textbf{0.959}$ & $\textbf{0.690}$ & $\textbf{0.580}$ \\ \hline
 \end{tabular}
\end{table}
\endgroup

\subsubsection{Confidence estimation on Librispeech}
We evaluated the performance of confidence estimation with different models (CEMs) on Librispeech, on the basis of the widely used metrics: the area under curve (AUC) of the receiver operating characteristic (ROC) curve and normalized cross entropy (NCE).
The ROC curve shows the false positive and true positive rates for different thresholds.
AUC values range from $0$ to $1$, and a higher AUC means a better estimator.
Let $\bm{c} = (c_1, ..., c_N)$ denote estimated confidence scores for all words, and $\bm{t} = (t_1, ..., t_N)$ denote their corresponding targets, where $t_i = 1$ for correct words and $0$ for incorrect words.
NCE is defined as
\begin{align}
NCE(\bm{c}, \bm{t}) = \frac{H(\bm{t}) - H(\bm{t}, \bm{c})}{H(\bm{t})}
\end{align}
where $H(\bm{t})$ denotes the entropy for the targets and $H(\bm{t}, \bm{c})$ denotes the binary cross entropy between the targets and estimated scores \cite{Siu99-EWC}.
NCE measures how close the confidence scores are to the targets and NCE $= 1$ for the perfect estimator.

The results are listed in Table \ref{tab:confidence}.
Note that the same models are used in rescoring and confidence estimation.
Confidence scores for ``ASR'' were obtained with the CTC forward-backward algorithm without any CEMs.
ELECTRA and P-ELECTRA provided good confidence scores even without fine-tuning.
By fine-tuning them, they outperformed fine-tuned BERT.
BERT was pre-trained to predict masked tokens, while ELECTRAs were already pre-trained to detect inappropriate tokens.
As shown in the bottom five rows of Table \ref{tab:confidence}, we investigated interpolating confidence scores of ASR and CEMs, as in Eq. (\ref{eq:score-confidence-interpolate}).
$\gamma$ was determined using the development sets.
For example, $\gamma = 0.6$ was suitable for fine-tuned P-ELECTRA.
By interpolation, we obtained far better confidence scores, which indicates CEMs pre-trained on large text corpora provide effective information that the ASR model does not provide.
Among them, P-ELECTRA achieved the best performance.

\subsubsection{Rescoring and confidence estimation on TED-LIUM2}
We also conducted rescoring and confidence estimation experiments on TED-LIUM2.
The results are listed in Table \ref{tab:tedlium2}.
They were evaluated on the ``test'' set, and the hyperparameters were adjusted using the ``dev'' set.
In rescoring, similar trends to Librispeech were observed, and fine-tuned P-ELECTRA reduced WER as much as BERT with faster inference.
In confidence estimation, CEMs themselves did not perform well, but the interpolation with ASR gave a large improvement.
Among the CEMs, fine-tuned P-ELECTRA performed the best.
On TED-LIUM2, the effect of fine-tuning was limited because of the smaller amount of paired data for generating the ASR $5$-best list compared with Librispeech.
Therefore, pre-training using phone information on text data was important for better performance.

\section{Conclusions}
We propose to apply ELECTRA to ASR rescoring and confidence estimation, in which ELECTRA detects ASR errors.
ELECTRA is pre-trained on large text corpora to predict whether each token is replaced by BERT or not.
However, there is a mismatch between ASR errors and token replacement by BERT.
Fine-tuning on ASR hypotheses can eliminate this mismatch, and we further propose phone-attentive ELECTRA to mitigate the mismatch also in pre-training on text.
In rescoring, we showed that ELECTRA was faster than Transformer LM because ELECTRA conducts binary classification with the sigmoid layer instead of the softmax computation, achieving competitive WER improvement.
In confidence estimation, we also showed that fine-tuned ELECTRA worked better than fine-tuned BERT, and the interpolation with the ASR confidence provided highly reliable confidence scores.
For future work, we will investigate corrupting inputs by not only replacement but also insertion and deletion in pre-training \cite{Wallach19-LT, Lewis20-BART} and predicting deletion errors \cite{Ragni18-CEDP} in rescoring.

\begingroup
\renewcommand{\arraystretch}{1.1}
\begin{table}[t]
  \label{tab:tedlium2}
  \caption{Rescoring and confidence estimation results on TED-LIUM2. ``AUC(+ASR)'' denotes AUC score by confidence score interpolation between ASR and CEM.}
  \vspace{10pt}
  \centering
  \begin{tabular}{lccc} \hline
      & Rescoring & \multicolumn{2}{c}{Confidence estimation} \\
      & WER $\downarrow$ & AUC $\uparrow$ & AUC(+ASR) $\uparrow$ \\ \hline
      ASR (CTC) & $12.48$ & $\textbf{0.914}$ & $0.914$ \\
      +Transformer LM & $9.90$ & $0.766$ & $0.914$ \\
      +BERT & $\textbf{9.83}$ & $0.854$ & $0.914$ \\
      +BERT (FT) & $10.47$ & $0.878$ & $0.939$ \\
      +ELECTRA & $10.03$ & $0.864$ & $0.930$ \\
      +ELECTRA (FT) & $10.00$ & $0.894$ & $0.939$ \\
      +P-ELECTRA & $9.89$ & $0.889$ & $0.937$ \\
      +P-ELECTRA (FT) & $\textbf{9.83}$ & $0.906$ & $\textbf{0.942}$ \\ \hline
      oracle & $7.52$ \\ \hline
  \end{tabular}
\end{table}
\endgroup

\bibliographystyle{IEEEbib}
\bibliography{strings,refs}

\end{document}